\documentclass[akbc,twoside,11pt,lettersize]{article}
\usepackage{akbc}
\usepackage{paralist}
\usepackage{tabularx}
\usepackage[caption = false]{subfig}
\usepackage[final]{graphicx}
\usepackage{epstopdf}
\akbcheading
\ShortHeadings{Cross-context News Corpus for Protest Events}{H\"{u}rriyeto\u{g}lu et al}
\finalcopy 
\begin{document}

\title{Cross-context News Corpus for Protest Events related Knowledge Base Construction}

\author{\name Ali H\"{u}rriyeto\u{g}lu \email ahurriyetoglu@ku.edu.tr \\
       \name Erdem Y\"{o}r\"{u}k \email eryoruk@ku.edu.tr \\
       \name Deniz Y\"{u}ret \email dyuret@ku.edu.tr \\
       \name Osman Mutlu \email omutlu@ku.edu.tr \\
       \name \c{C}a\u{g}r{\i} Yoltar \email cyoltar@ku.edu.tr \\
       \name F{\i}rat Duru{\c{s}}an \email fdurusan@ku.edu.tr \\
       \name Burak G\"{u}rel \email bgurel@ku.edu.tr \\
       \addr Rumelifeneri Yolu 34450 \\
       Sarıyer, İstanbul / Türkiye
    }
       

\maketitle

\begin{abstract}
We describe a gold standard corpus of protest events that comprise of various local and international sources from various countries in English. The corpus contains document, sentence, and token level annotations.
This corpus facilitates creating machine learning models that automatically classify news articles and extract protest event-related information, constructing knowledge bases which enable comparative social and political science studies.
For each news source, the annotation starts on random samples of news articles and continues with samples that are drawn using active learning. Each batch of samples was annotated by two social and political scientists, adjudicated by an annotation supervisor, and was improved by identifying annotation errors semi-automatically.
We found that the corpus has the variety and quality to develop and benchmark text classification and event extraction systems in a cross-context setting, which contributes to the generalizability and robustness of automated text processing systems.
This corpus and the reported results will set the currently lacking common ground in automated protest event collection studies.
\end{abstract}

\section{Introduction}

Socio-political event knowledge bases enable comparative social and political studies. The spatio-temporal distribution of these events sheds light on the causes and effects of governmental service management and political discourse resonate in the society. Consequently, construction of these knowledge bases a relatively long history. We focus on the protest events as a type of socio-political events. These events are in the scope of contentious politics (CP) and characterized by riots, strikes, and social movements, i.e. the ``repertoire of contention''~\cite{Giugni98,Tarrow94}.

Continuity of news articles and the significant impact of socio-political events direct social and political scientists to exploit news data to create knowledge bases of these events~\cite{Chenoweth+13,Weidman+19,Raleigh+10}. The need for collecting socio-political event data, which is protest and conflict data in the context of our work, has been satisfied manually~\cite{Yoruk12}, semi-automatically~\cite{Nardulli+15}, and automatically~\cite{Leetaru+13,Boschee+13,Schrodt+14,Sonmez+16}. Protest knowledge base creation is either too expensive for manual approaches or has serious limitations when it is performed automatically \cite{Wang+16,Ettinger+17,Ward+13}. Moreover, there has not been any common ground across projects that would enable to compare results of these studies~\cite{Lorenzini+16}. Therefore, as Emerging Welfare (EMW) project, we took on the challenge to create the common basis in terms of required high quality data and state-of-the-art tools for fully automating the creation of reliable and valid protest knowledge bases in a way that would serve as a benchmark and enable protest event collection studies benefit from it. This effort yielded a gold standard corpus (GSC) that will serve the machine learning and computational linguistics communities to study text processing tool development for constructing knowledge bases of protest events as well.

The GSC consists of English news articles from various local and international sources~\footnote{International sources were filtered based on meta-information to focus on the case countries.} from India, China, and South Africa. Variety of the sources has allowed studying cross-context robustness and generalizability, therefore addressing style and content change across sources, which are critical requirements of the ML models. The annotations were applied at document, sentence, and token level subsequently. Each instance is annotated by two people and adjudicated by the annotation supervisor for each level. Moreover, a detailed manual and semi-automatic quality check and error analysis were applied on the annotations.

The corpus contains more than 10,000 news articles labelled as protest and non-protest; protest containing articles, which are more than 800, were further labelled as containing event information or not at the sentence level, and annotated at the token level for detailed event information such as the trigger(s), semantic category, place, time and actor(s) of events, and co-reference information. The corpus has enabled the development of a pipeline of machine learning (ML) models that extract protest events from an archive of news articles. Moreover, some parts of the corpus has enabled shared-tasks that validated the corpus for cross-context document and sentence classification and token extraction~\cite{Hurriyetoglu+19a,Hurriyetoglu+19b}\footnote{\url{https://emw.ku.edu.tr/clef-protestnews-2019/}\textsuperscript{,}\url{https://emw.ku.edu.tr/?event=challenges-and-opportunities-in-automated-coding-of-contentious-political-events}} and event sentence coreference identification~\cite{Hurriyetoglu+20b}.

The contributions of this paper are
\begin{inparaenum}[1)]
\item a robust methodology for creating a corpus that enables the creation of robust ML-based text processing tools,
\item insights from applying this methodology to news archives across contexts,
\item a corpus that contains data from multiple contexts and annotations at various levels of granularity,
\item results of a pipeline consisting of automated tools that are created using the corpus, and
\item first state-of-the-art recall quantification on real random data in contrast to recall measurement in limited settings.
\end{inparaenum}

We will describe the context of our work in reference to recent related work in Section \ref{relwork}. Next we will introduce our methodology, the manuals we prepared, and the corpus we have created in Sections~\ref{methodology}, \ref{manual}, and~\ref{corpus}, respectively. Section~\ref{corpusevaluation} reports the results of the ML tools that were created using the corpus. Finally, Section~\ref{conclusion} concludes this report by drawing overall results and point to future steps we are planning to perform in the near future.


\section{Relevant work}
\label{relwork}
Automated socio-political event collection is shaped around language resources; automated tools that exploit these resources; the assumptions made to complete the design of an event collection system; and datasets that are output of these systems. The shared resources are rare and the accessible tools are a few. Moreover, the assumptions made in delivering a resulted dataset are not examined in diverse settings. 
Automated tools for event information collection are designed in terms of pipelines that receive news articles from one or more news sources and yield records of event information. Each tool is inherently limited to the language resources it facilitates and the setting it is validated. Therefore, the quality of the result when an automated tool is used to analyze different sources, i.e. cross-context performance of these pipelines is rarely evaluated. One of the first steps of these pipelines, which is discriminating between relevant and irrelevant documents, has been extensively reported in~\citet{Croicu+15} and~\citet{Hanna17}. Key term lists and labelled documents aid in determining which news reports contain relevant events. Each study reports their own key term lists and the way they use it. Moreover, labelled documents are released in terms of their URLs or document IDs in some collection without their content~\cite{Makarov+15}. Accessing the dataset using this limited information is the responsibility of the people who want to use these resources. Our work should be considered as a new version that does not restrict itself with keywords and apply state-of-the-art ML models to tackle the problem of selecting documents that contain event information, which is known as report selection problem in this field.

The second main step of this task is extracting relevant tokens. ACE~\cite{Doddington+2004} and TAC-KBP~\cite{Mitamura+2015} are language resources that can be exploitet for token level event extraction. The 
size and scope of the data for protest events provided by these resources is quite limited. Moreover, the event definition of ACE and TAC-KBP does not capture the CP events. For instance, the ACE definition of the event type DEMONSTRATE, in itself, is too restrictive to be applicable in terms of a broad understanding of CP for two reasons. First, as it seems to limit the scope of this event type to spontaneous (that is unorganized) gatherings of people, it excludes certain actions of political and/or grassroots organizations such as political parties and NGOs. Protest actions of such organizations sometimes do not involve mass participation despite aiming at challenging authorities, raising their political agendas or issuing certain demands. Putting up posters, distributing brochures, holding press declarations in public spaces are examples of such protest events. Secondly, the requirement of mass participation in a public area leaves many protest actions such as on-line mass petitions and boycotts, which are not necessarily tied to specific locations where people actually gather, and actions of individuals or small groups such as hunger strikes and self-immolation. Consequently, protest event specific annotation schemas and datasets were proposed~\cite{Huang+16,Sonmez+16,Makarov+16}. These kind of resources are mainly created using key terms for a single context and it is a challenge to obtain the datasets on the basis of the shared limited information. We follow the detailed protest event information tradition as proposed by \citet{Lorenzini+16} and \citet{Gerner+2002}, work on data not restricted with key terms, and make our data available to researchers in terms of shared-tasks and sufficient information to enable any researcher to work on it.

Recent studies often assume one or more of the following i) analyzing thousands or millions of sources will compensate for low tool performance, i.e. recall, ii) a news report contains information about a single event, iii) analyzing a sentence individually is sufficient to extract relevant information about an event, and iv) tool performance on a new source will be comparable to the performance on the validation setting~\cite{Weischedel+18,Tanev+08}. Quantifying effect of these assumptions is not a simple task, therefore they are rarely tested. We either provide our observations that shed light on the effect of these assumptions or do not make them.

\section{Methodology}
\label{methodology}

A gold standard corpus of protest events that can enable large scale, multi-source socio-political studies should be representative of the content it aims to capture. Moreover, it should enable quantifying the performance of the automation across contexts. Therefore, using available corpora that are already being allowed to be facilitated in research such as English Gigaword~\cite{Parker+11} is not an option for this setting. In order to satisfy these requirements, our methodology is designed to contain and incorporate multiple sources and countries and apply a detailed annotation methodology without sacrificing quality. 

We collect online local news articles, and international sources when local sources are not accessible, from India, China, and South Africa. We first download URLs of the freely accessible parts of online news archives such as Indian Express (IEX), New Indian Express (NIEX), The Hindu (TH), Times of India (ToI) and South China Morning Post (SCMP), People's Daily (PD).\footnote{The period covered by these archives is between around 2000 and 2017.} Then, for each source, we take a random sample of these URLs and download their content for labelling and annotation.\footnote{Only publicly accessible online information is processed and shared in terms of online URLs. We design our data collection, annotation, and tool development in a way that it would not yield any sensitive that could be used to target individuals by malicious state actors information about individuals. The precautions considered are: using and distributing data via URLs, and express personal characteristics in terms of broad categories such as student or worker.}

The random sampling approach makes the task challenging and closer to reality. Key term list causes some events such as the ones reported without using common terms of protest concept (the phrase ``classrooms empty'' can be used to report on a strike organized by teachers) to be missed~\cite{Hurriyetoglu+19a}. Moreover, lexical variance across contexts cannot always be captured using key terms. For instance, the terms ``bandh'' and ``idol immersion'' are event types that are specific to India and not covered by any general-purpose protest key terms list. Our evaluation of four key term lists, which are reported by \citet{Huang+16}, \citet{Wang+16}, \citet{Weidman+19}, and \citet{Makarov+16}, yielded .68 and .80 precision and recall on our randomly sampled batches at best.


Our annotation is based on an annotation manual created by an expert, randomly sampling documents from various sources and periods, and continuously monitoring the annotations to achieve a high inter-annotator agreement (IAA). The same manuals are applied on data collected from different sources and countries, which enables obtaining comparable measures of automatic tool performance across contexts. Finally, in order to eliminate the risk of wrong labelling due to lack of knowledge about a country, a domain expert in politics of the target country instructs the annotators before they start the annotation.

The annotation team consists of a supervisor, who is a social scientist and responsible for maintaining the annotation manuals and resolving the disagreements between the annotators, and master students or PhD candidates in social or political sciences, working in pairs. Throughout the annotation, the overlap ratio of annotated articles between pairs is 100\%. The annotation starts by labelling whether a news article mentions a protest. Then, the sentence(s) that contain protest information are identified. Finally, protest information such as participants, place, and time is detected in the protest-related sentences at token level. The three levels of annotation, are separate but integrated in the sense that they form a pipeline in which a single document goes through each individual step, and each step is built upon the result of the previous step, which is about completing a batch of documents for a specific level before we start to annotate at next level. The aim here is to maximize time and resource efficiency and performance by utilizing the feedback of each level of annotation for the whole process. The lack of clear boundaries between these levels at the beginning of the annotation project had caused a relatively lower IAA and more time to be spent on the quality check and correction of the dataset. For these same reasons, we add a new step, namely sentence level, to the aforementioned main steps of protest event pipelines. This order of tasks enables error analysis and optimization possibility during annotation and tool development efforts.


Each batch for document and sentence levels is corrected in terms of 
\begin{description}
\item[Spotchecks] \%10 of the agreements were checked by the annotation supervisor.
\item[ML-internal] 80\% of the batch is used to create an ML model. Next, the remaining 20\% is predicted using this model. This procedure was repeated until all instances are used at least once in training and once in test data.
\item[ML-external] The annotated and corrected data from previous batches is used to create an automated classifier that is used to classify the newly annotated batch.
\end{description} The disagreements between the classifiers and the annotations are checked manually for ML-internal and -external. Annotations that were checked based on spotcheck, ML-internal, and -external were found to be annotated wrongly at around 2\%, 50\%, and 10\% respectively. In total, around 10\% of the annotations were corrected using these measures. 

In the cases we need to improve the performance of a tool on an already covered source, adapt it to a recent period, or start the analysis of a new source, we apply a recall optimized active learning (AL) based sampling from a news archive. We first train multiple ML-based classifiers (three or more) on the available corpus, then predict a random batch from the new context. To achieve elevated recall scores, we take the logical-or of all classifiers as final prediction, and select positive samples to be annotated. Although the recall decrease from 100\% to 97\% in such a sample, the precision of it increase from around 5\% to around 70\% in comparison to a random sample.\footnote{This performance was measured on an AL sample and 200 news articles that were excluded from annotation at this sampling operation. The training data consisted of around 4,000 news articles that were randomly sampled and annotated from the same country of the resulted AL batch.} AL significantly decreases the effort the annotators should spend on annotation~\cite{Settles09}. Since annotators observe many more positive samples in this setting, we expect the decrease of recall to be a minor issue at the moment.

\section{Annotation manuals}
\label{manual}

Each annotation task has its respective annotation manual which defines the task and lays out the rules of annotation, which also enumerates certain cases which might be confusing and elaborates the context of a number of concrete situations and examples, for that task. \footnote{The annotation manuals can be found on \url{https://github.com/emerging-welfare/general_info/tree/master/annotation-manuals}.} The linguistic units of annotation, i.e. types of texts, for consecutive annotation tasks are documents, sentences, and expressions, i.e. groups of tokens respectively. 

\textbf{EMW Event Labeling Manual (ELAM)} is created for the document level annotation. This manual lays out the protest event ontology, that is, the protest event definition which specifies the range of CP events that are included in the scope of the project. Also, it contains the rules by which the news articles are decided to contain CP events. In a nutshell, CP events cover any politically motivated collective action which lays outside the official mechanisms of political participation associated with formal government institutions of the country in which the said action takes place. This broad event definition is developed and fleshed out in two sections. The first section identifies three abstract categories of collective action, namely, political mobilizations, social protests, and group confrontations, in order to define the broad range of socio-political events that the project simply refers to as protest events. Next, five specific categories of CP events are identified as concrete manifestations of the types of collective action already defined. Demonstrations, industrial actions, group clashes, political violence, armed militancy, and electoral mobilization events are the concrete types of events that the event ontology of the project encompasses. Once the event definition is laid out as such, certain criteria to which the news stories that report protest events must conform in order to be classified as protest news articles are enumerated. These criteria are the necessity of civilian actors, and the existence of concrete time and place information which ascertains that the event(s) the report mentions has definitely taken place. Only the news reports that mention events that have taken place in the past or are taking place at the time of writing are labeled as protest news articles. The references to the future (i.e. planned, threatened, announced or expected) events are not labeled, with the exception of threats of or attempts at violent events.\footnote{Although planned events and protest threats could have a role in our analysis~\cite{Huang+16}, they are neither relevant in the CP context nor their prevalence, which is below 0.5\% of a random sample according to our observations, allow their automated analysis.} 

\textbf{EMW Event Sentence Annotation Manual (ESAM)} enumerates rules according to which sentences of news reports are classified into protest event sentences and non-event sentences. Similar to the document level, sentences which contain references to the protest events are labeled as protest event sentences. The protest event sentences are defined as those sentences which give information about an event present in the news report and contain at least one direct reference to a protest event. That is to say, all event sentences must contain an expression which denotes the event. 

\textbf{EMW Information Extraction Annotation Manual (IEAM)} acts as the guide to the annotation on the token level. The IEAM defines all variables, pieces of information about protest events that the EMW project aims at extracting, and lays out the rules according to which expressions in the event sentences are to be annotated by using tags. There are general rules which apply to all tags, as well as specific rules which apply to individual tags. The tags and their corresponding variables are grouped under event, participant, organizer and target characteristics. Event characteristics are those variables that give information about the event itself, the main one being event trigger, i.e. event expressions which either directly denote the event or refer to it. The rest of the event characteristics and all other variables that are listed in participant, organizer, and target characteristics are event arguments. As it is mentioned above, event triggers are linguistic units which must exist in event sentences. All event arguments that exist within the text will only be annotated in event sentences. This is to ensure that arguments belong to their respective events unambiguously. Semantic categories of events are different types of collective action that the events in the articles fall into. Demonstrations (rallies, marches, sit-ins, slogan shouting, gatherings etc.), industrial actions (strikes, slow-downs, picket lines, gheraos etc.), group clashes (fights, clashes, lynching etc.), armed militancy (attacks, bombings, assassinations etc.) and electoral politics events (election rallies) are the event categories involved. Every event trigger is annotated as event type or mention, as well as one of the semantic categories that it is in. Participant and organizer characteristics contain participant type, name, ideology, religious, ethnic and caste identity, and socioeconomic status information about the actors which engage in protest. Participants are any individuals or groups who actively engage in the protest action, that is they are present at the event itself. Organizers are most commonly organizations that hold or take part in the protest events such as political parties, NGOs, unions etc. In some cases, influential individuals or leaders might be the organizers of the protest events. Individuals are annotated as organizers only in special cases where the article designates them explicitly as organizers or leaders of the protests. Target characteristics consist of target type and name and designates the antagonists of the protest events in the article, should they have one. The protests might target governments, officials, leaders, political organizations or other social groups in case of group clashes. 

Below are examples of how event information that is annotated on the token level. The bold tokens are the event triggers. The underlined tokens are event arguments, namely organizer name, participant count, participant type, participant caste, event time, event place, facility type, and target name. Note that the event which has not taken place, the rally, in the second sentence is not annotated.\footnote{We treat the event triggers and any other expressions that have a hyphen between them as a single token, e.g., `stone-pelting'. But, when there is not a hyphen between words, which is the case for `shouting slogans', the expression consists of two tokens. The first token is annotated as B-trigger and the following token(s) are annotated as I-trigger.} 

\begin{enumerate}
    \item 	\textbf{It} took a communal turn that had resulted in \textbf{stone-pelting}, \textbf{arson} and \textbf{loot}.
    \item The \underline{Bhim Army} and other \underline{Dalit} groups were refused permission to organize a rally against \textbf{atrocities} \underline{on May 9}, sparking off \textbf{violence} and \textbf{vandalism}, with several vehicles and buses \textbf{burnt}.
    \item \underline{At noon}, \underline{BJP} \underline{workers} \textbf{gathered} \underline{in the square} and \textbf{shouted slogans}, condemning the failure of the \underline{Union Government} in delivering justice to the victims of \underline{last year}'s terror \textbf{attack} \underline{at the train station} where armed \underline{militants} \textbf{killed} 25 people.
    \item \underline{In Bangalore}, \underline{hundreds} of \underline{workers} \textbf{participated} in the \textbf{rally} \underline{in front of the collectorate}.
\end{enumerate}

\section{Corpus}
\label{corpus}

We annotate the corpus in three levels that are i) \begin{inparadesc}
\item[Document] is what a reader see on a news article. It consists of a title, publication time, and the article text. ii)
\item[Sentence] is a text unit that ends with a sentence completing punctuation mark, e.g., period or question mark. iii)
\item[Token] is a punctuation mark or sequence of alphanumeric characters that is characterized as word in English. 
\end{inparadesc} The IAA for document and sentence levels are above .75 and .65 Krippendorf's alpha~\cite{Krippendorff+16} on average. The IAA for the token level is less consistent than other levels. Therefore, it is provided for each information type in Table~\ref{table:tokstats}. We interpret these scores as indication of how hard is the task and how the annotation manuals are able to enable the process.

The document counts for each document level batch are reported in Table~\ref{table:docstats}. Each batch is named after the source it was sampled from. In case we annotate some data from a corpus such as EventStatus~\cite{Huang+16} (ES)\footnote{\url{https://catalog.ldc.upenn.edu/LDC2017T09}, accessed on November 25th, 2019.} and RCV1~\cite{Lewis+04}\footnote{\url{https://trec.nist.gov/data/reuters/reuters.html}, accessed on November 25th, 2019.} that are readily released, we use their names as batch names. Suffixes were added to distinguish between different batches from the same source. For instance, SCMP1 and SCMP2 differ in terms of the period, which is 2000-2002 and 2000-2017 respectively, they cover.





\begin{table}[]
\begin{tabular}{|l|l|l|l|l|l|l|l|l|l|l|}
\hline
      & ES & INT & IEX & NIEX & PD & RCV1 & SCMP1 & SCMP2 & TH & ToI                  \\ \hline
Protest         & 151         & 262      & 296           & 71               & 69             & 802         & 17    & 19    & 264      & 481                           \\ \hline
Non-Protest     & 149         & 738      & 265           & 630              & 732            & 367         & 985   & 483   & 782      & 1985                          \\ \hline
Sampling & K           & AL       & AL            & R                & R              & AL          & R     & R     & AL       & \multicolumn{1}{l|}{R\&AL} \\ \cline{11-11} 
\hline
\end{tabular}
\caption{Document label statistics and sampling method. K, R, and AL indicate Key term, Random, and Active Learning respectively. ES, INT, and RCV1 are EventStatus, International, and Reuters which is filtered for China using meta-information.}
\label{table:docstats}
\end{table}

Active learning was applied to create three batches, which are INT2 (Guardian), SCMP3, and NIEX2 (New Indian Express), of articles that were annotated at the sentence level and reported in Table~\ref{table:sentstats}. The whole documents were sampled and their sentences were annotated. The high number of non-protest sentence annotations are caused by the documents that do not contain any protest information.





\begin{table}[!h]
\begin{center}
\begin{tabular}{|c|c|c|c|}

      \hline
       & INT2 & SCMP3 & NIEX2 \\
      \hline
      Protest & 1,658 & 511 & 1,299 \\ 
      \hline
      Non-protest & 9,045 & 2,847 & 7,083 \\ 
      \hline

\end{tabular}
\caption{Sentence level statistics. The total number of sentences and their annotations as protest and non-protest are reported.}
\label{table:sentstats}
\end{center}
\end{table}

Sentences of a subset of the positive documents were annotated at the token level. The number of the information types in the annotated documents, which are 704 and 135 from India and China respectively, are reported in the Table~\ref{table:tokstats}. A news article is annotated at the token level only if the event happens in the same country of the source or the country under focus for international sources, because protest event characteristics that are different across countries may affect the quality of the annotation.\footnote{Around 10\% of the positively annotated documents at the document level in a random sample reports a protest event that does not occur in a country under focus.} 

      


\begin{table}[]
\centering
\begin{tabular}{|c|c|c|c|c|c|c|c|}
\hline
Tag name                  & Time                       & Trigger                    & Place                      & Facility                   & Participant                & Organizer                  & Target                     \\ \hline
India                     & 822                        & 1,378                      & 645                        & 392                        & 2,283                      & 1,260                      & 1,453                      \\ \hline
China                     & 144                        & 142                        & 82                         & 52                         & 272                        & 88                         & 109                        \\ \hline
\multicolumn{1}{|c|}{IAA} & \multicolumn{1}{c|}{60.07} & \multicolumn{1}{c|}{50.02} & \multicolumn{1}{c|}{41.82} & \multicolumn{1}{c|}{39.10} & \multicolumn{1}{c|}{39.50} & \multicolumn{1}{c|}{47.44} & \multicolumn{1}{c|}{34.38} \\ \hline
\end{tabular}
\caption{Token level statistics and IAA in terms of Krippendorf's alpha}
\label{table:tokstats}
\end{table}

The separate event count per document is 1, 2, 3, 4, 5, 6 or more in 60\%, 23\%, 7\%, 5\%, 2\%, and 3\% of the documents respectively. Moreover, the distribution of the semantic event types demonstrations, industrial actions, group clashes, political violence and armed militancy, electoral mobilization, and other events is 55.5\%, 8.9\%, 13.7\%, 18.8\%, 2\%, 1.1\%. These numbers illustrate the first quantification of the multi-event and event type distribution phenomena in a random sample of news articles in protest domain and show that the assumption, made for instance by \citet{Tanev+08}, of a news article contain a single event inherently misses a significant amount of event information.


The document and sentence level annotations are stored in JSON and token level data is stored in FoLiA~\cite{Gompel+13} formats. We distribute the corpus in a way that does not violate copyright of the news sources. This involves only sharing information that is needed to reproduce the corpus from the source in cases it is not allowed to distribute the news articles. Namely, the document and sentence level data is downloaded using software we have developed and packaged in a Docker image. These software tools download, clean, and align text are provided in a Docker image in order to facilitate ease of use and reproducibility. The validation of this software was performed during the aforementioned shared-tasks.\footnote{Please follow the instructions on the Global Contentious Politics Gold Standard Data (GLOCON GOLD) repository to obtain the corpus: \url{https://github.com/emerging-welfare/glocongold}}

\section{Evaluation}
\label{corpusevaluation}

We have exploited the data from India in the corpus to train ML-based models using BERT~\cite{Devlin+19}, for document and sentence classification and token extraction in various scenarios. We fine-tune the pretrained BERT-Base with our data. Of each model for every level, hyperparameters are the same as original authors', except for our sentence classifier which restricts maximum sequence length to 128 instead of 512.\footnote{For our document model, if we split the text in subparts smaller than 512 tokens and take the logical or of each subpart's prediction as that document's prediction, the performance increases 2-3 F1 macro points in comparison to just using the first 512 tokens in a document for prediction.} Table~\ref{table:permodelperf} provides F1-macro scores of the document and sentence classification and F1-score that is based on CoNLL 2003 evaluation script for token extraction models for held-out data from India, for China and international\footnote{The international data is our Guardian sample that is filtered using active learning for China.} data in the corpus. The token extraction score is only the trigger detection performance in this table.

\begin{table}[!h]
\centering
\begin{tabular}{|c|c|c|c|c|}

      \hline
       & India & China & Int-China & South Africa \\
      \hline
      Document & .89 & .82 & .83 & .85\\ 
      \hline
      Sentence & .85 & .79 & .83 & .85 \\ 
      \hline
      Token & .74 & .67 & N/A & N/A \\ 
      \hline

\end{tabular}
\caption{F1-macro of document and sentence classification and F1 for trigger extraction.}
\label{table:permodelperf}
\end{table}

The token level scores, which are based on the BERT-base model fine-tuned on our GSC and are generated using a held-out part of it, are reported in Table~\ref{table:tokendetailed}. Additionally, we fine-tuned the Flair NER model \cite{Akbik+18}\footnote{\url{https://github.com/flairNLP/flair}, accessed on April 5, 2020.}, which is trained on CoNLL 2003 NER~\cite{Sang+03}, on our data by mapping our place, participant, and organizer tags to ``LOC'', ``PER'', and ``ORG'' in CoNLL data respectively. This model yielded significantly better results, which are .780, .697, and .652 for the place, participant, and organizer types respectively, in comparison to the BERT-base model. Finally, we run an event extraction model, which is again a BERT-base model, that is trained on ACE event extraction data on the same test data. We measured the trigger detection performance of this model based on its CONFLICT category predictions. The F1 scores of the CONFLICT type are .543 and .479 on its own and on our new data respectively. The difference between the scores  obtained using ACE and our training data show that our efforts contributes to the protest event collection studies significantly.


\begin{table}[]
\centering
\begin{tabular}{|c|c|c|c|c|c|c|c|}
\hline
          & Trigger & Time & Place & Facility & Participant & Organizer & Target \\ \hline
Precision & 0.756         & 0.663      & 0.724       & 0.436         & 0.649       & 0.568     & 0.497  \\ \hline
Recall    & 0.691         & 0.704      & 0.646       & 0.436         & 0.564       & 0.619     & 0.485  \\ \hline
F1        & 0.722         & 0.683      & 0.683       & 0.436         & 0.604       & 0.593     & 0.491  \\ \hline
\end{tabular}
\caption{Token level information extraction scores per information type.}
\label{table:tokendetailed}
\end{table}


We integrate the tools reported in Table~\ref{table:permodelperf} and report their performance on a separate 200 news articles dataset, which consists of 100 positively and 100 negatively predicted documents at document level, from India in Table~\ref{table:pipelinereview}.\footnote{We exclude 15 documents that contain events not related to India.} \textit{Doc}, \textit{Sent}, and \textit{Tok} correspond to the tool applied in the order the tool name is mentioned in the configuration name. The highest precision, recall, and F1-macro was yielded by \textit{Doc+Sent+Tok}, \textit{Tok}, and \textit{Doc+Tok} respectively. The event trigger detection score is the reported one for the \textit{Tok}. The performance of the trigger detection is lower than the one reported in Table~\ref{table:tokendetailed}, since this evaluation setting contain non-protest documents as well. The obvious result is that each additional component improve precision, but decrease the recall. The interesting result here is that integrating only the document classification tool enhance precision and a slight decrease in recall in comparison to other configurations.




Additional results that were yielded using some parts of this corpus can be found in various publications. The results obtained in a shared-task for the cross-context document and sentence classification and token extraction were reported in the overview paper of the ProtestNews Lab \cite{Hurriyetoglu+19b}, which was held in the scope of Conference and Labs of the Evaluation Forum (CLEF 2019). Moreover, the dataset was facilitated the event sentence coreference identification task, in which the participants developed systems to identify sentences about the same event in the scope of the workshop Automated Extraction of Socio-political Events from News (AESPEN) at Language Resources and Evaluation Conference (LREC 2020)~\cite{Hurriyetoglu+20b}. Participants of the shared tasks reported comparable results to the performance reported here.

\begin{table}[]
\centering
\begin{tabular}{|c|c|c|c|c|}
\hline
\multicolumn{1}{|c|}{} & Tok    & Sent+Tok & Doc+Tok       & Doc+Sent+Tok         \\   \hline
Precision   & .624                           & .696     & .660                           & \textbf{.701} \\ 
\hline
Recall  & \textbf{.663} & .561     & .647   & .547                           \\ 
\hline
F1      & .643  & .621     & \textbf{.653} & .614      \\ 
\hline
\end{tabular}
\caption{Trigger identification performances in various configurations of a pipeline.}
\label{table:pipelinereview}
\end{table}

\section{Conclusion and Future Work}
\label{conclusion}

We introduced a GSC that enables benchmarking and creation of automated tools for protest information collection across contexts. The methodology we have developed to ensure quality of the corpus, our observations during applying this methodology, and the results obtained using automated tools created using the corpus were reported in detail. The clear performance drop as the test data differ from training data, which is known as domain or covariate shift problem~\cite{Storkey+06}, shows how it is critical to incorporate cross-context aspects to the corpus and evaluation. Handling each source separately is the solution to improve reliability of the performance scores. Our ML models were created using training data from only a single country. Following steps should incorporate data from multiple contexts at model creation phase~\cite{Li+18,He+18}.

We keep track of what is included and excluded at each level in order to better automatize the task and allow quantification of the recall, which has been missing in this field. Restricting datasets by using key terms or basing a protest knowledge base on a subset of a source due to practical reasons was harming the validity and reliability of the resulted datasets. Starting with a random sample and continuing with recall optimized active learning during the creation of the gold standard corpus ensures training data will improve quality of the final gold standard dataset. 


We will extend the GSC with news sources in English, Portuguese, and Spanish and semantic categories such as violent vs. non-violent and urban vs. rural. Moreover, we will handle documents and sentences that contain multiple events. The recent models assume there is a separate event for each event trigger that is identified by the token extractor~\cite{Weischedel+18}. However, our observations directed us to identify and link the triggers that denote the same event~\cite{Ruppenhofer+10, Gabbard+11}. We will be developing tools for linking the event triggers about the same event in our pipeline~\cite{Lu+18}. 

\section*{Acknowledgements}
The authors are funded by the European Research Council (ERC) Starting Grant 714868 awarded to Dr. Erdem Y\"{o}r\"{u}k for his project Emerging Welfare.

\bibliographystyle{plainnat}
\bibliography{emwwp2}

\begin{thebibliography}{42}
\providecommand{\natexlab}[1]{#1}
\providecommand{\url}[1]{\texttt{#1}}
\expandafter\ifx\csname urlstyle\endcsname\relax
  \providecommand{\doi}[1]{doi: #1}\else
  \providecommand{\doi}{doi: \begingroup \urlstyle{rm}\Url}\fi

\bibitem[Akbik et~al.(2018)Akbik, Blythe, and Vollgraf]{Akbik+18}
Alan Akbik, Duncan Blythe, and Roland Vollgraf.
\newblock Contextual string embeddings for sequence labeling.
\newblock In \emph{Proceedings of the 27th International Conference on
  Computational Linguistics}, pages 1638--1649, Santa Fe, New Mexico, USA,
  August 2018. Association for Computational Linguistics.
\newblock URL \url{https://www.aclweb.org/anthology/C18-1139}.

\bibitem[Boschee et~al.(2013)Boschee, Natarajan, and Weischedel]{Boschee+13}
Elizabeth Boschee, Premkumar Natarajan, and Ralph Weischedel.
\newblock {Automatic Extraction of Events from Open Source Text for Predictive
  Forecasting}.
\newblock In V.S. Subrahmanian, editor, \emph{Handbook of Computational
  Approaches to Counterterrorism}, pages 51--67. Springer New York, New York,
  NY, 2013.
\newblock ISBN 978-1-4614-5311-6.
\newblock \doi{10.1007/978-1-4614-5311-6\_3}.
\newblock URL \url{https://doi.org/10.1007/978-1-4614-5311-6\_3}.

\bibitem[Chenoweth and Lewis(2013)]{Chenoweth+13}
Erica Chenoweth and Orion~A Lewis.
\newblock {Unpacking nonviolent campaigns: Introducing the NAVCO 2.0 dataset}.
\newblock \emph{Journal of Peace Research}, 50\penalty0 (3):\penalty0 415--423,
  2013.
\newblock \doi{10.1177/0022343312471551}.
\newblock URL \url{https://doi.org/10.1177/0022343312471551}.

\bibitem[Croicu and Weidmann(2015)]{Croicu+15}
Mihai Croicu and Nils~B Weidmann.
\newblock Improving the selection of news reports for event coding using
  ensemble classification.
\newblock \emph{Research \& Politics}, 2\penalty0 (4):\penalty0
  2053168015615596, 2015.
\newblock \doi{10.1177/2053168015615596}.
\newblock URL \url{https://doi.org/10.1177/2053168015615596}.

\bibitem[Devlin et~al.(2019)Devlin, Chang, Lee, and Toutanova]{Devlin+19}
Jacob Devlin, Ming-Wei Chang, Kenton Lee, and Kristina Toutanova.
\newblock {BERT}: Pre-training of deep bidirectional transformers for language
  understanding.
\newblock In \emph{Proceedings of the 2019 Conference of the North {A}merican
  Chapter of the Association for Computational Linguistics: Human Language
  Technologies, Volume 1 (Long and Short Papers)}, pages 4171--4186,
  Minneapolis, Minnesota, June 2019. Association for Computational Linguistics.
\newblock \doi{10.18653/v1/N19-1423}.
\newblock URL \url{https://www.aclweb.org/anthology/N19-1423}.

\bibitem[Doddington et~al.(2004)Doddington, Mitchell, Przybocki, Ramshaw,
  Strassel, and Weischedel]{Doddington+2004}
George Doddington, Alexis Mitchell, Mark Przybocki, Lance Ramshaw, Stephanie
  Strassel, and Ralph Weischedel.
\newblock The automatic content extraction ({ACE}) program {--} tasks, data,
  and evaluation.
\newblock In \emph{Proceedings of the Fourth International Conference on
  Language Resources and Evaluation ({LREC}{'}04)}, Lisbon, Portugal, May 2004.
  European Language Resources Association (ELRA).
\newblock URL \url{http://www.lrec-conf.org/proceedings/lrec2004/pdf/5.pdf}.

\bibitem[Ettinger et~al.(2017)Ettinger, Rao, Daum{\'e}~III, and
  Bender]{Ettinger+17}
Allyson Ettinger, Sudha Rao, Hal Daum{\'e}~III, and Emily~M. Bender.
\newblock {Towards Linguistically Generalizable NLP Systems: A Workshop and
  Shared Task}.
\newblock In \emph{Proceedings of the First Workshop on Building Linguistically
  Generalizable NLP Systems}, pages 1--10. Association for Computational
  Linguistics, 2017.
\newblock URL \url{http://aclweb.org/anthology/W17-5401}.

\bibitem[Gabbard et~al.(2011)Gabbard, Freedman, and Weischedel]{Gabbard+11}
Ryan Gabbard, Marjorie Freedman, and Ralph Weischedel.
\newblock Coreference for learning to extract relations: Yes virginia,
  coreference matters.
\newblock In \emph{Proceedings of the 49th Annual Meeting of the Association
  for Computational Linguistics: Human Language Technologies}, pages 288--293,
  Portland, Oregon, USA, June 2011. Association for Computational Linguistics.
\newblock URL \url{https://www.aclweb.org/anthology/P11-2050}.

\bibitem[Gerner et~al.(2002)Gerner, Schrodt, Yilmaz, and Abu-Jabr]{Gerner+2002}
Deborah~J Gerner, Philip~A Schrodt, Om{\"u}r Yilmaz, and Rajaa Abu-Jabr.
\newblock Conflict and mediation event observations (cameo): A new event data
  framework for the analysis of foreign policy interactions.
\newblock \emph{International Studies Association, New Orleans}, 2002.

\bibitem[Giugni(1998)]{Giugni98}
Marco~G. Giugni.
\newblock {Was It Worth the Effort? The Outcomes and Consequences of Social
  Movements}.
\newblock \emph{Annual Review of Sociology}, 24:\penalty0 371--393, 1998.
\newblock ISSN 03600572, 15452115.
\newblock URL \url{http://www.jstor.org/stable/223486}.

\bibitem[Hanna(2017)]{Hanna17}
Alex Hanna.
\newblock {MPEDS: Automating the Generation of Protest Event Data}, Jan 2017.
\newblock URL \url{osf.io/preprints/socarxiv/xuqmv}.

\bibitem[He et~al.(2018)He, Lee, Ng, and Dahlmeier]{He+18}
Ruidan He, Wee~Sun Lee, Hwee~Tou Ng, and Daniel Dahlmeier.
\newblock Adaptive semi-supervised learning for cross-domain sentiment
  classification.
\newblock In \emph{Proceedings of the 2018 Conference on Empirical Methods in
  Natural Language Processing}, pages 3467--3476, Brussels, Belgium,
  October-November 2018. Association for Computational Linguistics.
\newblock \doi{10.18653/v1/D18-1383}.
\newblock URL \url{https://www.aclweb.org/anthology/D18-1383}.

\bibitem[Huang et~al.(2016)Huang, Cases, Jurafsky, Condoravdi, and
  Riloff]{Huang+16}
Ruihong Huang, Ignacio Cases, Dan Jurafsky, Cleo Condoravdi, and Ellen Riloff.
\newblock Distinguishing past, on-going, and future events: The eventstatus
  corpus.
\newblock In \emph{Proceedings of the 2016 Conference on Empirical Methods in
  Natural Language Processing}, pages 44--54, 2016.

\bibitem[H{\"u}rriyeto{\u{g}}lu
  et~al.(2019{\natexlab{a}})H{\"u}rriyeto{\u{g}}lu, Y{\"o}r{\"u}k, Y{\"u}ret,
  Yoltar, G{\"u}rel, Duru{\c{s}}an, and Mutlu]{Hurriyetoglu+19a}
Ali H{\"u}rriyeto{\u{g}}lu, Erdem Y{\"o}r{\"u}k, Deniz Y{\"u}ret,
  {\c{C}}a{\u{g}}r{\i} Yoltar, Burak G{\"u}rel, F{\i}rat Duru{\c{s}}an, and
  Osman Mutlu.
\newblock A task set proposal for automatic protest information collection
  across multiple countries.
\newblock In Leif Azzopardi, Benno Stein, Norbert Fuhr, Philipp Mayr, Claudia
  Hauff, and Djoerd Hiemstra, editors, \emph{Advances in Information
  Retrieval}, pages 316--323, Cham, 2019{\natexlab{a}}. Springer International
  Publishing.
\newblock ISBN 978-3-030-15719-7.

\bibitem[H{\"u}rriyeto{\u{g}}lu
  et~al.(2019{\natexlab{b}})H{\"u}rriyeto{\u{g}}lu, Y{\"o}r{\"u}k, Y{\"u}ret,
  Yoltar, G{\"u}rel, Duru{\c{s}}an, Mutlu, and Akdemir]{Hurriyetoglu+19b}
Ali H{\"u}rriyeto{\u{g}}lu, Erdem Y{\"o}r{\"u}k, Deniz Y{\"u}ret,
  {\c{C}}a{\u{g}}r{\i} Yoltar, Burak G{\"u}rel, F{\i}rat Duru{\c{s}}an, Osman
  Mutlu, and Arda Akdemir.
\newblock Overview of clef 2019 lab protestnews: Extracting protests from news
  in a cross-context setting.
\newblock In Fabio Crestani, Martin Braschler, Jacques Savoy, Andreas Rauber,
  Henning M{\"u}ller, David~E. Losada, Gundula Heinatz~B{\"u}rki, Linda
  Cappellato, and Nicola Ferro, editors, \emph{Experimental IR Meets
  Multilinguality, Multimodality, and Interaction}, pages 425--432, Cham,
  2019{\natexlab{b}}. Springer International Publishing.
\newblock ISBN 978-3-030-28577-7.

\bibitem[H{\"u}rriyeto{\u{g}}lu et~al.(2020)H{\"u}rriyeto{\u{g}}lu, Zavarella,
  Tanev, Y{\"o}r{\"u}k, Safaya, and Mutlu]{Hurriyetoglu+20b}
Ali H{\"u}rriyeto{\u{g}}lu, Vanni Zavarella, Hristo Tanev, Erdem Y{\"o}r{\"u}k,
  Ali Safaya, and Osman Mutlu.
\newblock Automated extraction of socio-political events from news ({AESPEN}):
  Workshop and shared task report.
\newblock In \emph{Proceedings of the Workshop on Automated Extraction of
  Socio-political Events from News 2020}, pages 1--6, Marseille, France, May
  2020. European Language Resources Association (ELRA).
\newblock ISBN 979-10-95546-50-4.
\newblock URL \url{https://www.aclweb.org/anthology/2020.aespen-1.1}.

\bibitem[Krippendorff et~al.(2016)Krippendorff, Mathet, Bouvry, and
  Widl{\"o}cher]{Krippendorff+16}
Klaus Krippendorff, Yann Mathet, St{\'e}phane Bouvry, and Antoine
  Widl{\"o}cher.
\newblock On the reliability of unitizing textual continua: Further
  developments.
\newblock \emph{Quality and quantity}, 50\penalty0 (6):\penalty0 2347--2364,
  2016.

\bibitem[Leetaru and Schrodt(2013)]{Leetaru+13}
Kalev Leetaru and Philip~A Schrodt.
\newblock {GDELT: Global data on events, location, and tone, 1979--2012}.
\newblock In \emph{ISA annual convention}, volume~2, pages 1--49. Citeseer,
  2013.

\bibitem[Lewis et~al.(2004)Lewis, Yang, Rose, and Li]{Lewis+04}
David~D. Lewis, Yiming Yang, Tony~G. Rose, and Fan Li.
\newblock Rcv1: A new benchmark collection for text categorization research.
\newblock \emph{J. Mach. Learn. Res.}, 5:\penalty0 361--397, December 2004.
\newblock ISSN 1532-4435.
\newblock URL \url{http://dl.acm.org/citation.cfm?id=1005332.1005345}.

\bibitem[Li et~al.(2018)Li, Yang, Song, and Hospedales]{Li+18}
Da~Li, Yongxin Yang, Yi-Zhe Song, and Timothy Hospedales.
\newblock {Learning to Generalize: Meta-Learning for Domain Generalization}.
\newblock In \emph{AAAI Conference on Artificial Intelligence}, 2018.
\newblock URL
  \url{https://www.aaai.org/ocs/index.php/AAAI/AAAI18/paper/view/16067/16547}.

\bibitem[Lorenzini et~al.(2016)Lorenzini, Makarov, Kriesi, and
  Wueest]{Lorenzini+16}
Jasmine Lorenzini, Peter Makarov, Hanspeter Kriesi, and Bruno Wueest.
\newblock {Towards a Dataset of Automatically Coded Protest Events from
  English-language Newswire Documents}.
\newblock In \emph{Paper presented at the Amsterdam Text Analysis Conference},
  2016.

\bibitem[Lu and Ng(2018)]{Lu+18}
Jing Lu and Vincent Ng.
\newblock Event coreference resolution: A survey of two decades of research.
\newblock In \emph{Proceedings of the Twenty-Seventh International Joint
  Conference on Artificial Intelligence, {IJCAI-18}}, pages 5479--5486.
  International Joint Conferences on Artificial Intelligence Organization, 7
  2018.
\newblock \doi{10.24963/ijcai.2018/773}.
\newblock URL \url{https://doi.org/10.24963/ijcai.2018/773}.

\bibitem[Makarov et~al.(2015)Makarov, Lorenzini, Rothenh{\"a}usler, and
  W{\"u}est]{Makarov+15}
Peter Makarov, Jasmine Lorenzini, Klaus Rothenh{\"a}usler, and Bruno W{\"u}est.
\newblock Towards automated protest event analysis.
\newblock In \emph{New Frontiers of Automated Content Analysis in the Social
  Sciences}, July 2015.
\newblock URL \url{https://doi.org/10.5167/uzh-143877}.

\bibitem[Makarov et~al.(2016)Makarov, Lorenzini, and Kriesi]{Makarov+16}
Peter Makarov, Jasmine Lorenzini, and Hanspeter Kriesi.
\newblock Constructing an annotated corpus for protest event mining.
\newblock In \emph{Proceedings of the First Workshop on {NLP} and Computational
  Social Science}, pages 102--107, Austin, Texas, November 2016. Association
  for Computational Linguistics.
\newblock \doi{10.18653/v1/W16-5613}.
\newblock URL \url{https://www.aclweb.org/anthology/W16-5613}.

\bibitem[Mitamura et~al.(2015)Mitamura, Liu, and Hovy]{Mitamura+2015}
Teruko Mitamura, Zhengzhong Liu, and Eduard~H. Hovy.
\newblock Overview of {TAC} {KBP} 2015 event nugget track.
\newblock In \emph{Proceedings of the 2015 Text Analysis Conference, {TAC}
  2015, Gaithersburg, Maryland, USA, November 16-17, 2015, 2015}. {NIST}, 2015.
\newblock URL
  \url{https://tac.nist.gov/publications/2015/additional.papers/TAC2015.KBP\_Event\_Nugget\_overview.proceedings.pdf}.

\bibitem[Nardulli et~al.(2015)Nardulli, Althaus, and Hayes]{Nardulli+15}
Peter~F. Nardulli, Scott~L. Althaus, and Matthew Hayes.
\newblock {A Progressive Supervised-learning Approach to Generating Rich Civil
  Strife Data}.
\newblock \emph{Sociological Methodology}, 45\penalty0 (1):\penalty0 148--183,
  2015.
\newblock \doi{10.1177/0081175015581378}.
\newblock URL \url{https://doi.org/10.1177/0081175015581378}.

\bibitem[Parker et~al.(2011)Parker, Graff, Kong, Chen, and Maeda]{Parker+11}
Robert Parker, David Graff, Junbo Kong, Ke~Chen, and Kazuaki Maeda.
\newblock {English Gigaword Fifth Edition.}, 2011.
\newblock URL \url{https://catalog.ldc.upenn.edu/LDC2011T07}.

\bibitem[Raleigh et~al.(2010)Raleigh, Linke, Hegre, and Karlsen]{Raleigh+10}
Clionadh Raleigh, Andrew Linke, H{\aa}vard Hegre, and Joakim Karlsen.
\newblock Introducing acled: an armed conflict location and event dataset:
  special data feature.
\newblock \emph{Journal of peace research}, 47\penalty0 (5):\penalty0 651--660,
  2010.

\bibitem[Ruppenhofer et~al.(2010)Ruppenhofer, Sporleder, Morante, Baker, and
  Palmer]{Ruppenhofer+10}
Josef Ruppenhofer, Caroline Sporleder, Roser Morante, Collin Baker, and Martha
  Palmer.
\newblock Semeval-2010 task 10: Linking events and their participants in
  discourse.
\newblock In \emph{Proceedings of the 5th International Workshop on Semantic
  Evaluation}, SemEval '10, pages 45--50, Stroudsburg, PA, USA, 2010.
  Association for Computational Linguistics.
\newblock URL \url{http://dl.acm.org/citation.cfm?id=1859664.1859672}.

\bibitem[Schrodt et~al.(2014)Schrodt, Beieler, and Idris]{Schrodt+14}
Philip~A Schrodt, John Beieler, and Muhammed Idris.
\newblock {Three'sa charm?: Open event data coding with el: Diablo, Petrarch,
  and the open event data alliance}.
\newblock In \emph{ISA Annual Convention}, 2014.

\bibitem[Settles(2009)]{Settles09}
Burr Settles.
\newblock Active learning literature survey.
\newblock Computer Sciences Technical Report 1648, University of
  Wisconsin--Madison, 2009.

\bibitem[S{\"o}nmez et~al.(2016)S{\"o}nmez, {\"O}zg{\"u}r, and
  Y{\"o}r{\"u}k]{Sonmez+16}
{\c{C}}a{\u{g}}{\i}l S{\"o}nmez, Arzucan {\"O}zg{\"u}r, and Erdem
  Y{\"o}r{\"u}k.
\newblock Towards building a political protest database to explain changes in
  the welfare state.
\newblock In \emph{Proceedings of the 10th SIGHUM Workshop on Language
  Technology for Cultural Heritage, Social Sciences, and Humanities}, pages
  106--110. Association for Computational Linguistics, 2016.
\newblock \doi{10.18653/v1/W16-2113}.
\newblock URL \url{http://www.aclweb.org/anthology/W16-2113}.

\bibitem[Storkey and Sugiyama(2006)]{Storkey+06}
Amos~J Storkey and Masashi Sugiyama.
\newblock Mixture regression for covariate shift.
\newblock In \emph{Proceedings of the 19th International Conference on Neural
  Information Processing Systems}, NIPS'06, pages 1337--1344, Cambridge, MA,
  USA, 2006. MIT Press.
\newblock URL \url{http://dl.acm.org/citation.cfm?id=2976456.2976624}.

\bibitem[Tanev et~al.(2008)Tanev, Piskorski, and Atkinson]{Tanev+08}
Hristo Tanev, Jakub Piskorski, and Martin Atkinson.
\newblock Real-time news event extraction for global crisis monitoring.
\newblock In Epaminondas Kapetanios, Vijayan Sugumaran, and Myra Spiliopoulou,
  editors, \emph{Natural Language and Information Systems}, pages 207--218,
  Berlin, Heidelberg, 2008. Springer Berlin Heidelberg.
\newblock ISBN 978-3-540-69858-6.

\bibitem[Tarrow(1994)]{Tarrow94}
S.~Tarrow.
\newblock \emph{{Power in Movement: Social Movements, Collective Action and
  Politics}}.
\newblock Cambridge Studies in Comparative Politics. Cambridge University
  Press, 1994.
\newblock ISBN 9780521422710.
\newblock URL \url{https://books.google.com.tr/books?id=hN5nQgAACAAJ}.

\bibitem[Tjong Kim~Sang and De~Meulder(2003)]{Sang+03}
Erik~F. Tjong Kim~Sang and Fien De~Meulder.
\newblock Introduction to the conll-2003 shared task: Language-independent
  named entity recognition.
\newblock In \emph{Proceedings of the Seventh Conference on Natural Language
  Learning at HLT-NAACL 2003 - Volume 4}, CONLL ’03, page 142–147, USA,
  2003. Association for Computational Linguistics.
\newblock \doi{10.3115/1119176.1119195}.
\newblock URL \url{https://doi.org/10.3115/1119176.1119195}.

\bibitem[van Gompel and Reynaert(2013)]{Gompel+13}
Maarten van Gompel and Martin Reynaert.
\newblock Folia: A practical xml format for linguistic annotation - a
  descriptive and comparative study.
\newblock \emph{Computational Linguistics in the Netherlands Journal},
  3:\penalty0 63--81, Dec. 2013.
\newblock URL \url{https://clinjournal.org/clinj/article/view/26}.

\bibitem[Wang et~al.(2016)Wang, Kennedy, Lazer, and Ramakrishnan]{Wang+16}
Wei Wang, Ryan Kennedy, David Lazer, and Naren Ramakrishnan.
\newblock {Growing pains for global monitoring of societal events}.
\newblock \emph{Science}, 353\penalty0 (6307):\penalty0 1502--1503, 2016.
\newblock ISSN 0036-8075.
\newblock \doi{10.1126/science.aaf6758}.
\newblock URL \url{http://science.sciencemag.org/content/353/6307/1502}.

\bibitem[Ward et~al.(2013)Ward, Beger, Cutler, Dickenson, Dorff, and
  Radford]{Ward+13}
Michael~D Ward, Andreas Beger, Josh Cutler, Matthew Dickenson, Cassy Dorff, and
  Ben Radford.
\newblock Comparing gdelt and icews event data.
\newblock \emph{Event Data Analysis}, 21\penalty0 (1):\penalty0 267--297, 2013.

\bibitem[Weidmann and R{\o}d(2019)]{Weidman+19}
Nils~B. Weidmann and Espen~Geelmuyden R{\o}d.
\newblock \emph{The Internet and Political Protest in Autocracies}, chapter
  Coding Protest Events in Autocracies.
\newblock Oxford Studies in Digital Politics, Oxford, 2019.

\bibitem[Weischedel and Boschee(2018)]{Weischedel+18}
Ralph Weischedel and Elizabeth Boschee.
\newblock What can be accomplished with the state of the art in information
  extraction? a personal view.
\newblock \emph{Comput. Linguist.}, 44\penalty0 (4):\penalty0 651–658,
  December 2018.
\newblock ISSN 0891-2017.
\newblock \doi{10.1162/coli_a_00331}.
\newblock URL \url{https://doi.org/10.1162/coli_a_00331}.

\bibitem[Yoruk(2012)]{Yoruk12}
Erdem Yoruk.
\newblock \emph{The politics of the Turkish welfare system transformation in
  the neoliberal era: Welfare as mobilization and containment}.
\newblock The Johns Hopkins University, 2012.

\end{thebibliography}

\end{document}